# Large Scale Strongly Supervised Ensemble Metric Learning, with Applications to Face Verification and Retrieval

Chang Huang
NEC Laboratories America
chuang@sv.nec-labs.com

Shenghuo Zhu
NEC Laboratories America
zsh@sv.nec-labs.com

Kai Yu
NEC Laboratories America
kyu@sv.nec-labs.com

## Abstract

*Learning Mahanalobis distance metrics in a high-dimensional feature space is very difficult especially when structural sparsity and low rank are enforced to improve computational efficiency in testing phase. This paper addresses both aspects by an ensemble metric learning approach that consists of sparse block diagonal metric ensembling and joint metric learning as two consecutive steps. The former step pursues a highly sparse block diagonal metric by selecting effective feature groups while the latter one further exploits correlations between selected feature groups to obtain an accurate and low rank metric. Our algorithm considers all pairwise or triplet constraints generated from training samples with explicit class labels, and possesses good scalability with respect to increasing feature dimensionality and growing data volumes. Its applications to face verification and retrieval outperform existing state-of-the-art methods in accuracy while retaining high efficiency.*

## 1. Introduction

Similarity measurement has been studied intensively in fields such as machine learning, information retrieval, artificial intelligence, and cognitive science for a long time; it plays a crucial role in learning, reasoning and predicting as similar things usually behave similarly [10]. Consider machine learning algorithms such as K-means, nearest-neighbors classifiers and kernel methods – their performance critically relies on being given a good metric over the input space [23]. Metric learning aims at finding appropriate similarity measurements between any pair of instances that preserve desired distance structure. This is not only fundamental to understanding high level concepts such as categories, but also necessary for many low level tasks such as verification, clustering and retrieval. Especially in recent years, many supervised metric learning algorithms have been proposed to learn Mahanalobis distance metrics for clustering or $k$-nearest neighbor classification, which can be divided into two categories according to supervision types. The first category is *weakly supervised* which learns metrics from directly provided pairwise constraints between instances. Examples are Xing et al.'s pioneering distance metric learning method [23] and Davis et al.'s Information-Theoretic Metric Learning (ITML) [3]. Such weak constraints are also known as *side information* [23]. The second category is *strongly supervised* which requires explicit class labels assigned to every instance and generates a potentially large number of constraints between them. It includes Globerson and Roweis's Metric Learning by Collapsing Classes (MCML) [7] and Weinberger et al.'s Large Margin Nearest Neighbor (LMNN) [22]. These supervised methods generally perform well in data sets with up to hundreds of features, but are still very limited in tasks with high dimensional data. Usually, the dimensionality is reduced by Principle Component Analysis (PCA) beforehand, which can allow noise to overwhelm signal that is useful for *supervised* metric learning as PCA is essentially an *unsupervised* metric learning method. This problem becomes even more serious when using overcomplete representations of data where huge redundancy needs to be addressed carefully.

Overcomplete representations of data possess great robustness in the presence of noise and other forms of degradations and thus, are better suited to subsequent processing [17]. In this paper, we propose a *strongly supervised ensemble metric learning* approach for *low-rank* Mahanalobis distance metrics based on a *sparse combination* of features from an overcomplete set. It consists of two consecutive steps: *sparse block diagonal metric ensembling* and *joint metric learning*. The former step sequentially selects effective features and learns their associated weak metrics that correspond to diagonal blocks of Mahanalobis matrix in the entire feature space. The latter step learns another Mahanalobis distance metric in the feature subspace enabled by the former step, by jointly considering already selected features with an optional low-rank constraint, so as to represent all instances in an even lower dimensional space. This two-step approach can be viewed as a supervised sparse dimensionality reduction followed by supervised metric learning.

Unlike previous metric learning methods, our implementation adopts a convex smooth loss function based on an exponential logit surrogate function. We also develop an efficient batch learning algorithm for this strongly supervised approach to enable simultaneously handling of all pairwise or triplet constraints generated from a large amount of data with explicit class labels.

We validate our approach on the Label Faces in the Wild (*LFW*) data set [13] for face verification and another much larger data set for face retrieval. In the unrestricted configuration of *LFW*, our approach improves the mean classification accuracy from the previous record $91.30\%$ (by a commercial system from face.com [19]) to $92.58\%$ without using any outside data, heuristic knowledge or 3D face models. For face retrieval in the extended data set, our approach encodes each face image into a highly discriminative $150$-dimensional vector. This retains high retrieval accuracy, while achieving significantly lower computational cost compared to state-of-the-art (such as LMNN [22]).

The contributions of this paper are three-fold:

- We introduce the group sparsity into metric learning problems with overcomplete representations of data.
- We develop a two-step batch learning algorithm for efficient learning of such metrics in large scale problems.
- Our algorithm considerably improves the best performance in the unrestricted configuration of LFW.

## 2. Related Work

We limit the discussion of related work to supervised metric learning, where supervision is induced by a set of similar/dissimilar constraints between instances. For weakly supervised metric learning, Xing et al. [23] formulate metric learning as a constrained convex programming problem that minimizes distance between within-class instances with the constraint that between-class instances are apart from each other by a certain distance. Relevant Components Analysis (RCA) [2] learns a global linear transformation from the similar constraints, which is improved in Discriminative Component Analysis (DCA) [12] by exploiting dissimilar constraints; both are extensions of Linear Discriminant Analysis (LDA) [4]. Information-Theoretic Metric Learning (ITML) [3] formulates the metric learning as a Bregman optimization problem by minimizing LogDet divergence between the learned metric and the prior metric subject to similar/disimilar constraints. As for strongly supervised cases, Neighborhood Component Analysis (NCA) [8] learns a distance metric by extending the nearest neighbor classifier, and the Large-Margin Nearest Neighbor (LMNN) [22] fits this idea into a maximum margin framework. Metric Learning by Collapsing Classes (MCML) [7] tries to find a Mahalanobis distance that can collapse within-class instances onto a single point. A much more comprehensive survey of metric learning methods can be found in [24].

Of all previous work on metric learning, [18] and [15] share a relatively similar sparsification idea to our proposed method. They both derive LogDet divergence objective function with element-wise $\ell^1$ regularization from ITML [3]; the former one tries to solve the dual problem by block coordinate descent algorithm [6] while the latter one directly addresses the primal problem by alternating linearization method [9]. However, as each feature corresponds to a row and a column in the Mahanalobis distance matrix, such element-wise regularization is not suitable for sparsifying feature selections that prefer matrices with whole empty rows and columns; whereas straightforwardly applying group lasso with row-wise and column-wise $\ell^1$ regularization is usually very expensive in high dimensional feature space. Instead, the first step of our approach progressively constructs a small collection of effective features during metric learning procedure, which can be considered as an analogy of AdaBoost [5] or matching pursuit [16] with group $\ell^0$ regularization.

## 3. Supervised Ensemble Metric Learning

We start this section with symbols and notions.

- An instance is represented by $K$ feature groups as

$$x = [x^{(1)}, x^{(2)}, \cdots, x^{(K)}]^\top \in \mathbb{R}^D,\ x^{(k)} \in \mathbb{R}^d,$$

where $x^{(k)}$ is the $k$-th feature group with $d$ features and the concatenated feature dimensionality $D = Kd$.

- A *squared Mahanalobis distance metric* is

$$d_{ij}^A = (x_i - x_j)^\top A(x_i - x_j), \forall x_i, x_j \in \mathbb{R}^D, A \succeq 0,$$

where $A$ is a *Mahanalobis matrix*.

- $\mathcal{B} \subseteq \mathbb{R}^{D \times D}$ is the block matrix space in which matrices consist of $K \times K$ blocks, each of size $d \times d$.
- $\mathcal{B}_{kl}$ is the sparse block matrix space where only the block in the $k$-th row and the $l$-th column is non-zero.
- $\lfloor A \rfloor_{kl}$ is the projection of matrix $A$ onto space $\mathcal{B}_{kl}$.
- $\|A\|_F$, $tr(A)$ and $r(A)$ are Frobenius norm, trace norm and rank of $A$.
- $\|A\|_{s0} = \text{card}\{k | \lfloor A \rfloor_{kl} \neq 0 \vee \lfloor A \rfloor_{lk} \neq 0\ \exists l\}$ is the number of feature groups used by $A$, *i.e.*, our defined structural $\ell^0$ norm of $A$.
- $\Pi_{\text{PSD}}(A)$ is the projection of $A$ onto Positive Semi-Definite space; $\Pi_\nu(A)$ is to project eigenvalues of $A$ onto a simplex to make its trace norm lower than $\nu$.
- $\mathcal{X}$ is a training set with class labels for every sample.
- $x_i \sim x_j$ or $\pi_{ij} = +1$ denote $x_i$ and $x_j$ are of the same category; $x_i \not\sim x_k$ or $\pi_{ik} = -1$ denote they are of different categories.

- $N = |\mathcal{X}|$, $N_i^+ = |\{x_j \mid x_j \sim x_i, x_j \in \mathcal{X}\}|$ and $N_i^- = |\{x_k \mid x_k \not\sim x_i, x_k \in \mathcal{X}\}|$ are the total number of training samples, the number of within-class and the number of between-class samples to $x_i$.

## 3.1. Problem Formulation

In this paper, we discuss the situation that instances are represented by a large collection of fixed-size feature groups, without loss of generality to cases with varying-size feature groups. These feature groups could be subspaces of raw features, or wavelet descriptors at different positions and scales such as SIFT and LBP features. As there is huge redundancy in an overcomplete representation, a desired metric should avoid using feature groups with little discriminability so as to estimate similarities between instances efficiently without sacrifice in accuracy. For this purpose, we formulate the metric learning problem as

$$\min_A \quad f(A|\mathcal{X}) = \frac{\lambda}{2}\|A\|_F^2 + \ell(A|\mathcal{X}) \quad (1)$$
$$\text{subject to} \quad A \succeq 0, \|A\|_{s0} \leq \mu, \; tr(A) \leq \nu,$$

in which $\ell(A|\mathcal{X})$ is the *empirical loss function* regarding to discriminability of $A$ upon training set $\mathcal{X}$. The *regularization term* penalizes matrix $A$ by its squared Frobenius norm with coefficient $\lambda$ for better generalization ability; $A \succeq 0$ is to keep the learned metric satisfying triangle inequality; $tr(A) \leq \nu$ is to obtain a low-rank matrix $A$ so that every instance eventually can be represented in a low-dimensional space; and in particular, $\|A\|_{s0} \leq \mu$ is to impose group sparsity on matrix $A$ to insure that only a limited number of feature groups (smaller than $\mu$) will be actually involved in the testing phase.

## 3.2. A Two-Step Approach

However, the optimization task in Equation 1 is NP hard due to the structural $\ell^0$ norm and thus extremely difficult to solve with high-dimensional overcomplete representations of data. We propose a two-step metric learning approach consisting of *sparse block diagonal metric ensembling* and *joint metric learning* to address this problem, which are elaborated in Algorithm 1 and Algorithm 2 respectively.

Sparse block diagonal metric ensembling starts from an empty set of feature groups ($A = 0$), progressively chooses effective feature groups (indicated by $\kappa$), learns weak metrics ($A_\kappa^*$) and combines them into a strong one. Every candidate feature group is evaluated by the partial derivative of loss function $f(\cdot)$ w.r.t its corresponding diagonal block in matrix $A$. The opposite of this partial derivative matrix is projected onto Positive Semi-Definite space so that it decreases the loss function while keeping the updated matrix Positive Semi-Definite. The algorithm selects a diagonal block with the largest $\ell^2$ norm of its projected partial derivative matrix, and optimizes it as well as a scale factor $\alpha$ adjusting the previously learned matrix to minimize the loss function. After $\mu$ rounds of such weak metric learning procedure, we can obtain an sparse block diagonal matrix, $A_\dagger$, with at most $\mu$ feature groups activated. Through the eigenvalue decomposition, its orthogonal linear transformation matrix $L_\dagger$ preliminarily reduces the feature dimensionality from $D$ to $D_\dagger$ ($D_\dagger = r(A_\dagger) \ll D$).

**Algorithm 1** Sparse block diagonal metric ensembling
1: **INPUT**: $\mathcal{X}$, $\mu$ and $\lambda$.
2: $A \leftarrow 0$
3: **for** $t = 1$ **to** $\mu$ **do**
4: $\quad \kappa = \underset{k \in \{1,2,\cdots,K\}}{\operatorname{argmax}} \left\| \Pi_{\text{PSD}}\big(-\lfloor \frac{\partial f(A|\mathcal{X})}{\partial A} \rfloor_{kk}\big) \right\|_2$
5: $\quad A_\kappa^*, \alpha^* = \underset{A_\kappa \succeq 0, A_\kappa \in \mathcal{B}_{kk}, \alpha \in \mathbb{R}^+}{\operatorname{argmin}} f(\alpha A + A_\kappa | \mathcal{X})$
6: $\quad A \leftarrow \alpha^* A + A_\kappa^*$
7: **end for**
8: $A_\dagger \doteq A$, $L_\dagger \doteq U$ where $U \Lambda U^\top = A$, $U \in \mathbb{R}^{D \times D_\dagger}$.
9: **OUTPUT**: $A_\dagger$ and $L_\dagger$.

**Algorithm 2** Joint metric learning
1: **INPUT**: $\mathcal{X}$, $\nu$, $\lambda$ and $U_\dagger$.
2: Dimension reduction: $\mathcal{X}_\dagger = \{U_\dagger^\top x \mid x \in \mathcal{X}\}$.
3: $A \leftarrow 0$.
4: **while** not converge **do**
5: $\quad \nabla f(A|\mathcal{X}_\dagger) = \frac{\partial f(A|\mathcal{X}_\dagger)}{\partial A}$.
6: $\quad$ Choose a proper step $\gamma$.
7: $\quad A \leftarrow \Pi_\nu\big(A - \gamma \nabla f(A|\mathcal{X}_\dagger)\big)$.
8: **end while**
9: $L_\ddagger \doteq L_\dagger L$ where $LL^\top = A$.
10: $A_\ddagger \doteq L_\ddagger L_\ddagger^\top$.
11: **OUTPUT**: $A_\ddagger$ and $L_\ddagger$.

Owning to the supervised dimension reduction achieved by sparse block diagonal metric ensembling, the joint metric learning is capable of further exploiting correlations between those selected feature groups in the intermediate feature space $\mathcal{X}_\dagger$ without diagonal block constraints. The projected gradient descent method is adopted to solve this optimization problem: the Mahanalobis matrix is iteratively updated by its gradient with a proper step size, and then regulated by projecting its eigenvalues onto a simplex for satisfying $tr(A) \leq \nu$ and $A \succeq 0$. In this way, a secondary linear transformation matrix $L$ can be learned to map instances onto an even lower-dimensional space, and the final linear transformation matrix $L_\ddagger = L_\dagger L$ helps represent all instances in a $D_\ddagger$-dimensional space, where Euclidean distance is the optimal metric for similarity measurement. In other words, $A_\ddagger = L_\ddagger L_\ddagger^\top$ is the final Mahanalobis matrix.

The key component of this metric learning approach is the computation of empirical loss function $\ell(A|\mathcal{X})$ and its gradient, which is defined by constraints between instances.

From training data with explicit class labels, two types of constraints can be generated: pairwise and triplet. For example, let $x_i$ and $x_j$ be two instances of the same category and $x_k$ be the instance of another category. From the view point of $x_i$, on the one hand, pairwise constraints are $d_{ij}^A < \theta$ and $d_{ik}^A > \theta$ where $\theta$ is a general threshold separating all similar pairs from dissimilar ones; constraints of this type are adopted in verification problems that determines whether a pair of instances belong to the same category or not. On the other hand, the triplet constraint is $d_{ij}^A \leq d_{ik}^A$, which is clearly a ranking preference designed for clustering or retrieval tasks that only concern about relative difference of distances between instances. We discuss the two constraint types respectively in coming subsections.

### 3.3. Pairwise Constraints

The *empirical error* of $A$ with threshold $\theta$ on all pairwise constraints from $\mathcal{X}$ is defined by

$$
\begin{aligned}
\epsilon_\theta(A \mid \mathcal{X}) &= \Pr(\pi_{ij}(d_{ij}^A - \theta) > 0 \mid x_i, x_j \in \mathcal{X}) \quad (2) \\
&= \mathbb{E}_{x_i, x_j \in \mathcal{X}} \mathbf{1}_{\pi_{ij}(d_{ij}^A - \theta) > 0},
\end{aligned}
$$

in which $\pi_{ij} = \pm 1$ indicates whether $x_i$ and $x_j$ belong to the same category or not, and $\mathbf{1}_{(\cdot)}$ is a characteristic function that outputs 1 if $(\cdot)$ is satisfied or 0 otherwise. By replacing $\mathbf{1}_{(\cdot)}$ with the *exponential-based logit surrogate function* $\psi_\beta(e^z) \doteq \frac{\ln(1+\beta e^z)}{\ln(1+\beta)}$ and setting $\beta = 1$, we obtain an upper bound of this empirical error as

$$
\begin{aligned}
\epsilon_\theta(A \mid \mathcal{X}) &\leq \mathbb{E}_{x_i, x_j \in \mathcal{X}} \psi_1(e^{d_{ij}^A - \theta}) \quad (3) \\
&= \frac{1}{N^2 \ln 2} \sum_{i,j} \ln(1 + e^{\pi_{ij}(d_{ij}^A - \theta)}) \\
&\doteq \ell_\theta(A \mid \mathcal{X}),
\end{aligned}
$$

which is smooth and convex, serving as the *empirical loss function* with pairwise constraints.

Let $\eta_{ij} = d_{ij}^A - \theta$. By the chain rule, we have

$$
\begin{aligned}
\frac{\partial \ell_\theta(A \mid \mathcal{X})}{\partial A} &= \sum_{i,j} \frac{\partial \ell_\theta(A \mid \mathcal{X})}{\partial \eta_{ij}} \cdot \frac{\partial \eta_{ij}}{\partial A} \quad (4) \\
&= \sum_{i,j} w_{ij} \cdot (x_i - x_j)(x_i - x_j)^\top,
\end{aligned}
$$

in which the weight term

$$
w_{ij} \doteq \frac{\partial \ell_\theta(A \mid \mathcal{X})}{\partial \eta_{ij}} = \frac{\pi_{ij}}{N^2 \ln 2} \cdot \frac{e^{\pi_{ij}(d_{ij}^A - \theta)}}{1 + e^{\pi_{ij}(d_{ij}^A - \theta)}}. \quad (5)
$$

Given the weight matrix $W = \{w_{ij}\}_{N \times N}$, Equation 4 can be efficiently computed by

$$
\frac{\partial \ell_\theta(A \mid \mathcal{X})}{\partial A} = X(S - W - W^\top)X^\top, \quad (6)
$$

where $X = [x_1, x_2, \cdots, x_N]$ is the feature matrix of $\mathcal{X}$ and $S = \text{diag}(\sum_k w_{1k} + w_{k1}, \cdots, \sum_k w_{Nk} + w_{kN})$.

### 3.4. Triplet Constraints

The *empirical error* of $A$ on all triplet constraints from $\mathcal{X}$ is defined by

$$
\begin{aligned}
\epsilon(A \mid \mathcal{X}) &= \Pr(d_{ij}^A > d_{ik}^A \mid x_j \sim x_i, x_k \nsim x_i)) \quad (7) \\
&= \mathbb{E}_{x_i, x_j \sim x_i, x_k \nsim x_i} \mathbf{1}_{d_{ij}^A > d_{ik}^A}
\end{aligned}
$$

Similarly, we have an upper bound of this empirical error as

$$
\epsilon(A \mid \mathcal{X}) \leq \mathbb{E}_{x_i, x_j \sim x_i, x_k \nsim x_i} \psi_\beta(e^{d_{ij}^A - d_{ik}^A}) \doteq \overline{\ell}(A \mid \mathcal{X}). \quad (8)
$$

However, this is not an appropriate loss function as the computational complexity given $\{d_{ij}^A | \forall i, j\}$ could be $O(N^3)$. By using the concavity of $\psi_\beta(\cdot)$, we further relax it to be

$$
\begin{aligned}
\overline{\ell}(A \mid \mathcal{X}) &\leq \mathbb{E}_{x_i} \psi_\beta\left(\mathbb{E}_{x_j \sim x_i, x_k \nsim x_i} e^{d_{ij}^A - e_{ik}^A}\right) \quad (9) \\
&= \mathbb{E}_{x_i} \psi_\beta\left(\mathbb{E}_{x_j \sim x_i} e^{d_{ij}^A} \cdot \mathbb{E}_{x_k \nsim x_i} e^{-d_{ik}^A}\right) \\
&= \mathbb{E}_{x_i} \psi_\beta(\phi_i^+ \cdot \phi_i^-) \\
&\doteq \ell(A \mid \mathcal{X}),
\end{aligned}
$$

which is still smooth and convex, where

$$
\begin{aligned}
\phi_i^+ &= \mathbb{E}_{x_j \sim x_i} e^{d_{ij}^A} = \frac{1}{N_i^+} \sum_{x_j \sim x_i} e^{d_{ij}^A}, \quad (10) \\
\phi_i^- &= \mathbb{E}_{x_k \nsim x_i} e^{-d_{ik}^A} = \frac{1}{N_i^-} \sum_{x_k \nsim x_i} e^{-d_{ik}^A}.
\end{aligned}
$$

This is a loss function holding the upper bound of empirical error with all triplet constraints generated from $\mathcal{X}$, and its computational complexity given $\{d_{ij}^A | \forall i, j\}$ is just $O(N^2)$, the same as that with pairwise constraints in Equation 3.

Similar to Equation 4 and 5, we have

$$
\frac{\partial \ell(A \mid \mathcal{X})}{\partial A} = \sum_{i,j} w_{ij} \cdot (x_i - x_j)(x_i - x_j)^\top, \quad (11)
$$

$$
w_{ij} = \begin{cases} \frac{\beta \phi_i^+ \exp(d_{ij}^A)}{N N_i^+ \ln(1+\beta) \cdot (1 + \beta \phi_i^+ \phi_i^-)} & \text{if } x_j \sim x_i, \\ -\frac{\beta \phi_i^- \exp(-d_{ij}^A)}{N N_i^- \ln(1+\beta) \cdot (1 + \beta \phi_i^+ \phi_i^-)} & \text{if } x_j \nsim x_i, \end{cases}
$$

### 3.5. Computational Complexity

Let $T$ be the number of iterations needed for projected gradient descent. For both pairwise constraints and triplet constraints, sparse block diagonal metric ensembling costs $O(ND + N^2)$ in memory and $O(\mu(K+T)(N^2 d + Nd^2 + d^3))$ in time, which scales up well on a larger $K$, the number of feature groups. Joint metric learning costs $O(ND_\dagger)$ in memory and $O(T(N^2 D_\dagger + N D_\dagger^2 + D_\dagger^3))$ in time, which can afford more training data compared to sparse block diagonal metric ensembling as $D_\dagger \ll D = Kd$.

## 4. Experiments

We implement the proposed ensemble metric learning method by Python and C++ for face verification and face retrieval. To preprocess the data, every face image is aligned by its eyes and mouth, and then cropped to keep its central $110 \times 150$ region; a Gaussian smooth filter with 1-pixel kernel size is applied afterward to suppress white noise.

Two types of features are employed in our experiments: covariance matrix descriptors (CMD) [21] and soft local binary pattern histograms (SLBP) [1]. Covariance matrix descriptors represent image regions by the covariance matrix of basic features such as spatial location, intensity, higher order derivatives, etc. We follow the work in [21] where every covariance matrix descriptor is a $45$-dimensional vector. SLBP features are derived from local binary pattern histograms (LBP) by considering probabilities of a patch being different patterns, so each SLBP descriptor is a $59$-dimensional vector. In our experiments, 13,260 rectangular regions of size varying from $8 \times 8$ to $96 \times 144$ are enumerated within the $110 \times 150$ region, covering the face image homogeneously. Each rectangular region defines a covariance matrix descriptor and a SLBP descriptor; every descriptor is whitened on training data by PCA; top $45$ dimensions of SLBP descriptors are preserved. Eventually, each face image is represented by 13,260 covariance matrix descriptors and 13,260 SLBP descriptors, each of $45$ dimensions constituting a feature group. Both types of features do not use color information of the images. The flowchart of feature extraction and dimensionality reduction is illustrated in Figure 1.

### 4.1. Face Verification

The goal of face verification is to determine whether a pair of testing face images belong to the same person or not, which fits into the pairwise constraint formulation in Section 3.3. Our face verification experiment is carried out on the (*LFW*) data set [13], which is a very challenging data set with 13,233 face images of 5,749 persons collected from the web. It has two views for its own data. View 1 partitions the 5,749 persons into training and testing set for development purposes such as model selection and parameter tuning. View 2 divides them into ten disjoint folds and specifies 300 "positive" pairs and 300 "negative" pairs within each fold: positive pairs are instances of the same person while negative pairs are those of different persons. It is suggested in [13] to conduct a ten-fold cross validation experiment, in which models are required to be learned on nine folds and tested on the 600 pairs of the remaining fold in terms of classification accuracy. We follow the "unrestricted configuration" [13] that allows us to use the identity information of training data and choose the "aligned" version [20] of *LFW* for fair comparison to previous work on this data set.

Three different metrics are learned for testing on each fold: a CMD based one, a SLBP based one and a combined one by applying joint metric learning on concatenated intermediate features (*i.e.*, $\mathcal{X}_\dagger$ in Algorithm 2) of both types. We set $\mu = 400$ so that at most $400$ feature groups are selected; $\nu = \infty$ as representing face images in a low-dimensional space is not crucial for face verification; $\theta = 14$ and $\lambda = 1$ according to experiments on view 1. To alleviate the difficulty due to view point changes, we define the flip-free distance between $x_i$ and $x_j$ by

$$\overline{d^A}(x_i, x_j) = 0.25\big(d^A(x_i, x_j) + d^A(x'_i, x_j) \\ + d^A(x_i, x'_j) + d^A(x'_i, x'_j)\big),$$

where $x'_i$ and $x'_j$ are horizontally flipped images. Whether $x_i$ and $x_j$ are of the same person is determined upon this distance measurement with a threshold chosen on View 1.

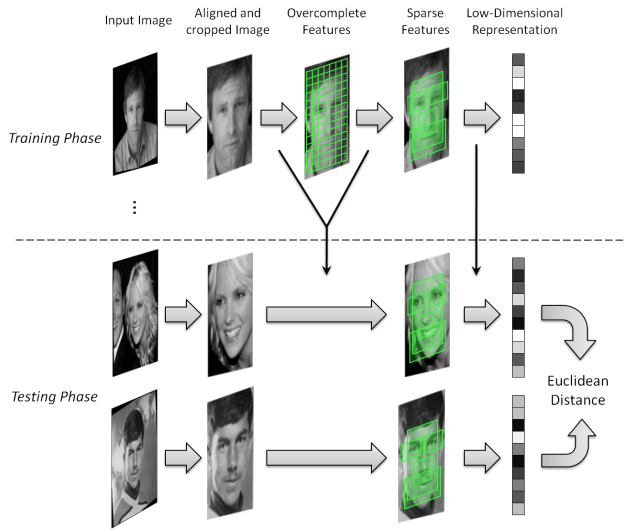

Figure 1. Flowchart of ensemble metric learning. The learning phase needs to generate overcomplete representations of face images while the testing phase only needs to extract a sparse collection of them according to the result of sparse block diagonal metric ensembling. Face images are finally coded by low-dimensional vectors and the Euclidean distance measures their similarities.

|   | CMD | SLBP | CMD+SLBP |
|---|---|---|---|
| $D$ | 596,700 | 596,700 | - |
| $r(A_\dagger)$ | $1627 \pm 53$ | $1507 \pm 34$ | - |
| $\bar{\epsilon}(A_\dagger)$ | $0.116 \pm 0.013$ | $0.145 \pm 0.02$ | - |
| $r(A_\ddagger)$ | $1312 \pm 15$ | $1179 \pm 10$ | $2302 \pm 21$ |
| $\bar{\epsilon}(A_\ddagger)$ | $0.083 \pm 0.011$ | $0.100 \pm 0.013$ | $0.074 \pm 0.014$ |

Table 1. Performance of the two steps of our approach on the two types of features and the combined one. Note the significant reduction in rank $r(A_\dagger)$ and testing error $\bar{\epsilon}(A_\dagger)$ of metrics learned by sparse block diagonal metric ensembling. Also note the further reduction in $r(A_\ddagger)$ and improvement in $\bar{\epsilon}(A_\ddagger)$, enabled by joint metric learning.

As shown in Table 1, from $D$ to $r(A_\dagger)$ (*i.e.*, $D_\dagger$), the di-mensionality of overcomplete representations is dramatically reduced by sparse block diagonal metric ensembling since only 400 features are adopted among the 13,260 ones and their corresponding diagonal blocks are not full rank due to the positive semi-definite constraint. From $r(A_\dagger)$ to $r(A_\ddagger)$, joint metric learning further decreases the dimensionality needed to represent each instance, even without the low-rank constraint ($\nu = \infty$ in this experiment), while reducing the testing error remarkably from $\bar{\epsilon}(A_\dagger)$ to $\bar{\epsilon}(A_\ddagger)$. Metrics based on CMD are overall superior to those based on SLBP, while better metrics are learned based on concatenated intermediate features of both types.

|  | mean $\pm$ std |
| --- | --- |
| LDML-MkNN, funneled [11] | $87.50 \pm 0.40\%$ |
| LBP multishot, aligned [20] | $85.17 \pm 0.61\%$ |
| combined multishot, aligned [20] | $89.50 \pm 0.51\%$ |
| LBP PLDA, aligned [14] | $87.33 \pm 0.55\%$ |
| combined PLDA, funneled&aligned [14] | $90.07 \pm 0.51\%$ |
| face.com r2011b[19] [19] | $91.30 \pm 0.30\%$ |
| CMD ensemble metric learning, aligned | $91.70 \pm 1.10\%$ |
| SLBP ensemble metric learning, aligned | $90.00 \pm 1.33\%$ |
| CMD + SLBP, aligned | $\mathbf{92.58 \pm 1.36}\%$ |

Table 2. Classification accuracy in the unrestricted configuration on *LFW*. With either a single type or multiple types of features, our proposed approach outperforms previous methods in both cases.

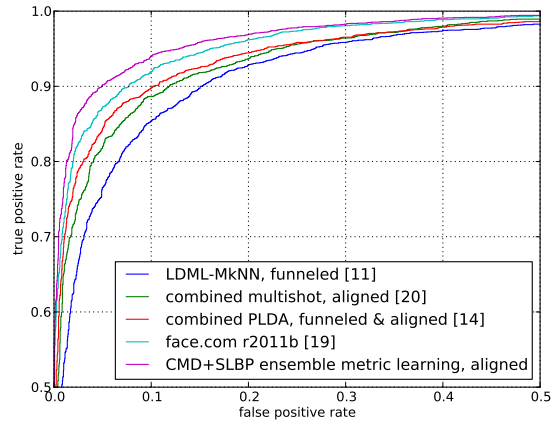

Figure 2. ROC curves of state-of-the-art methods using multiple types of features in unrestricted configuration of *LFW*.

Table 2 compares our ensemble metric learning approach to existing state-of-the-art methods in the same unrestricted configuration on *LFW*, including LDML-MkNN [11] that combines Logistic Discriminant based Metric Learning and Marginalized k-NN classifiers, multishot [20] that uses multiple one shot similarity score based on Information Theoretic Metric Learning [3], PLDA [14] that adopts Probabilistic Linear Discriminate Analysis instead of distance based methods, and face.com [19], a commercial system without any details published. With single type of features, the CMD based metrics achieve $91.70\%$ and the SLBP based ones achieve $90.00\%$ in average accuracy, both outperforming the best of previously published work ($87.33\%$ by LBP PLDA [14]); with multiple types of features, the ensemble metric learning method further improves the average accuracy to $92.58\%$, which is even better than the result of face.com [19] that is equipped by an accurate 3D face model to overcome pose and illumination variations. ROC curves of methods with multiple types of features are shown in Figure 2.

A data-driven method like ours does not require heuristics such as the face structure. Hence, our approach can easily generalize to metric learning problems on object types other than the face. But it also has the drawback of being sensitive to the choice of training data when there are only a limited amount of them available. This explains the larger standard deviation of our approach's classification accuracy compared to the other methods in Table 2. This drawback can be alleviated by using more training data.

Our method takes about one day to train a metric for one fold on five Intel 2.2GHz Xeon servers. The learned metric is of high computational efficiency owing to the group sparsity of features. In practice, it takes less than 50 ms to obtain the final vector representation of a face image on one server, and less than 1 ms to compute the Euclidean distance between two such vectors.

### 4.2. Face Retrieval

Unlike face verification, face retrieval aims at searching a reference database for registered faces that are similar to a given query. The triplet constraint formulation in Section 3.4 is adopted since only relative preference is considered in this situation. Three data sets are involved in this face retrieval experiment: *LFWext*, *subLFWext* and *Qry60*. *LFWext* includes 171,509 images of 4,740 persons, collected from internet based on the name list of *LFW*; *subLFWext* is a subset of *LFWext* containing 1,275 people, each having exactly 40 images; *Qry60* is a different set consisting of 2,040 images from 60 people outside the name list of LFW. Metrics are learned on *subLFWext* or *LFWext* and tested on *Qry60*. Only covariance matrix descriptor features are employed in this experiment for simplicity.

We study the proposed joint metric learning method with two benchmarks: LDA [4] and LMNN [22]. Their performance is evaluated by leave-one-out 3-NN classification error on *Qry60*, mean average precision of retrieving the same person with each of its images inside *Qry60*, and mean average precision by injecting *Qry60* into *LFWext* and retrieving them back. We denote these three evaluation metrics by $\epsilon_Q$, mAP$_Q$ and mAP$_{QL}$ hereafter.

As the first step, an intermediate metric $A_\dagger$ is learned by sparse block diagonal metric ensembling on *subLFWext*,

| $k$ | $\epsilon_Q$ | mAP$_Q$ | mAP$_{QL}$ | $r(A_\ddagger)$ |
|---|---|---|---|---|
| 5 | 0.0696 | 0.6563 | 0.3321 | 1033 |
| 10 | 0.0588 | 0.6926 | 0.3801 | 816 |
| 15 | 0.0578 | 0.7050 | 0.3975 | 714 |
| 20 | 0.0554 | 0.7125 | 0.4078 | 649 |
| 25 | 0.0544 | 0.7175 | 0.4134 | 602 |
| 30 | 0.0544 | 0.7207 | **0.4161** | 572 |
| 35 | **0.0539** | **0.7209** | 0.4155 | **561** |
| 39 | 0.0583 | 0.7093 | 0.3971 | 594 |

Table 3. Performance of joint metric learning with different $\overline{k}$ ($\overline{k} \leq 39$ as every person has exactly 40 face images in *subLFWext*).

with parameter settings as $\mu = 400$, $\beta = 10^6$ and $\lambda = 1$. The rank of $A_\dagger$ is 2513 so all face images are represented in a 2513-dimensional intermediate feature space.

### 4.2.1 Target neighbor number $\overline{k}$

The concept of target neighbors is introduced by LMNN [22] specifically for $k$-NN classification problems. Given an instance, among all instances of the same category, only the top $\overline{k}$ closest ones are adopted to generate triplet constraints for metric learning. It helps relax metric learning by eliminating too difficult triplet constraints that contain dissimilar but within-class instance pairs. Our approach can easily accommodate this concept by ignoring within-class instances that are not among the top $\overline{k}$ closest ones to $x_i$ during computing $\phi_i^+$ in Equation 10. We apply joint metric learning with different target neighbor numbers on *subLFWext*, and compare their results in terms of classification error, retrieval accuracy and the rank of final metrics. As shown in Table 3, optimal choices of $\overline{k}$ are either 30 or 35. On the one hand, targeting more neighbors can enforce more triplet constraints to make the learned metric generalize better to unknown testing categories like the 60 persons in *Qry60*. On the other hand, ignoring a few too difficult within-class pairs actually alleviates the risk of overfitting since they are very likely to be just outliers or even mislabeled data.

### 4.2.2 Joint metric learning vs. LMNN

The joint metric learning method with triplet constraints is an analogy to LMNN [22] in the nature of maximizing the margin between instances of different categories. The two methods differ in loss functions (smooth vs. hinge) and regularization terms. We reduce the dimensionality of the intermediate feature space to 100, 200, 300, 500 and 1000 by PCA, and set the target neighbor number $\overline{k}$ to be 5, 10 and 30 respectively to compare joint metric learning to the improved LMNN [1] implemented by the active set method. For joint metric learning we use $\nu = +\infty$, $\lambda = 1$, $\beta = 10^6$. For LMNN we use default parameters. We still use *subLFWext*

[1] LMNN2 @ http://www.cse.wustl.edu/ kilian/code/code.html

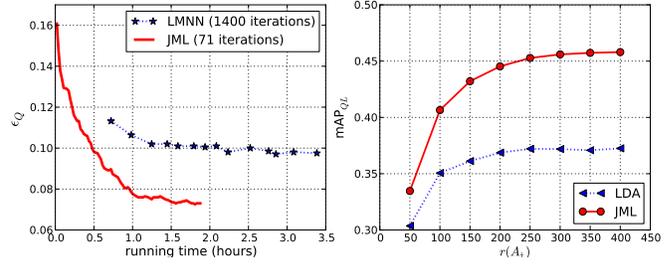

Figure 3. The left figure compares LMNN and joint metric learning (JML) in convergence speed of $\epsilon_Q$ with $\overline{k} = 5$ in 100 dimensional space; the right one compares LDA and joint metric learning (JML) in mAP$_{QL}$ with different final dimensionality.

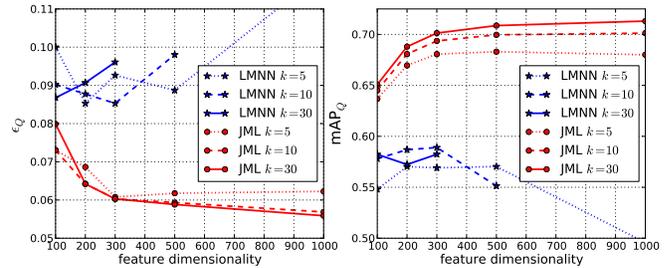

Figure 4. Comparisons between LMNN and joint metric learning (JML) in $\epsilon_Q$ and mAP$_Q$. Note the consistent improvement achieved by JML using more target neighbors and higher feature dimensionality.

for training. Experiments are done on a single Intel 2.2GHz Xero server.

As shown in the left of Figure 3, the leave-one-out 3-NN classification error by joint metric learning drops faster than that by LMNN in the 100-dimensional feature space when $\overline{k} = 5$; the former converges after 71 iterations while the latter runs for about 1,400 iterations. Owing to the relaxation of empirical loss with triplet constraints in Equation 9, joint metric learning is able to compute the exact gradient of the loss function efficiently. Though each iteration is relatively slow, it generally requires much fewer iterations to converge compared to the LMNN algorithm implemented by active set method. Especially for hard problems where different categories are not well separated, the active set has to maintain a huge number of triplet constraints and spends a significant amount of time in updating them. We set the maximum iteration number to be 1000 for LMNN, as it barely improves after that. According to comparisons in Figure 4, joint metric learning consistently benefits from more target neighbors and higher feature dimensionality, while LMNN seems unable to take advantage of these changes and even suffers from them.

From the time complexity reported in Table 4, LMNN does not scale well with respect to more target neighbors and higher-dimensional feature spaces. Metric learning by LMNN was not even feasible in three cases because the program exceeded the memory limit (the server has 16G memory). In contrast, our method can easily accommodate problems of larger scales and yields steady training times per iteration with any target neighbor number.

| | LMNN | | | Joint metric learning | | |
|---|---|---|---|---|---|---|
| dim | $\bar{k}=5$ | $\bar{k}=10$ | $\bar{k}=30$ | $\bar{k}=5$ | $\bar{k}=10$ | $\bar{k}=30$ |
| 100 | 2.5 | 3.5 | 6.4 | 1.9(71) | 1.6(60) | 1.2(45) |
| 200 | 4.3 | 5.8 | 13.8 | 6.0(190) | 2.6(80) | 1.9(59) |
| 300 | 6.9 | 9.7 | 28.7 | 4.9(119) | 3.8(99) | 2.6(70) |
| 500 | 15.1 | 25.1 | - | 7.1(130) | 6.6(119) | 4.4(81) |
| 1000 | 63.7 | - | - | 10.3(126) | 10.5(130) | 7.6(91) |

Table 4. Running hours of LMNN and joint metric learning. In the bracket are the numbers of iterations (1000 for LMNN). Three cases of LMNN failed in our experiment as they exceeded the memory limit of our server (16G).

### 4.3. Joint metric learning vs. LDA

In this subsection, we aim at the best practice of metric learning in the 2513-dimensional intermediate feature space on the *LFWext* data set that contains 171,509 images from 4,740 persons. Parameter settings are $\nu=6$, $\lambda=1$ and $\beta=10^6$. It takes about 27 hours to train the metric on 8 Intel 2.2GHz Xero servers. Though the rank of final metric $A_\ddagger$ is 800 owing to the trace norm constraint, it is still not low enough for face retrieval tasks. Additionally, we retain only $\bar{d}$ eigenvectors with the largest eigenvalues (*i.e.*, $r(A_\ddagger)=\bar{d}$) to represent every instance more economically. As LMNN cannot be applied to such a large-scale problem, we compare joint metric learning to LDA in their performance of retrieving *Qry60* from *LFWext + Qry60* (*i.e.*, mAP$_{QL}$). As shown in the right of Figure 3, joint metric learning significantly outperforms LDA that becomes saturated when $r(A_\ddagger)>200$. By comparing to the mAP$_{QL}$ score obtained by the metric learned on *subLFWext* (Table 3), we also observe considerable improvement in retrieval accuracy by using more training data.

In practice, we keep the top 150 projection vectors to represent a face by a 150-dimensional vector, which makes retrieval in large databases extremely efficient. On a single Intel Xero 2.2GHz server, it only takes about 2 seconds to exhaustively explore an unlabeled database with 4 million faces and find the most similar ones to a query.

## 5. Conclusion

This paper presents an efficient two-step metric learning algorithm for large scale problems with overcomplete representations of data. A highly sparse Mahanalobis distance metric is learned which selects only a small portion of effective feature groups. With this metric, every instance can be represented by a compact vector for efficient verification or retrieval. Our future work focuses on two aspects: 1) further relaxation of empirical loss function to pursue lower complexity for larger scale problems; 2) a regularization method that can incorporate prior knowledge easily.

## References

[1] T. Ahonen and M. Pietikainen. Soft histograms for local binary patterns. *Finnish Signal Processing Symposium*, 2007.
[2] A. Bar-Hillel, T. Hertz, N. Shental, and D. Weinshall. Learning distance functions using equivalence relations. *ICML*, 2003.
[3] J. V. Davis, B. Kulis, P. Jain, S. Sra, and I. S. Dhillon. Information-theoretic metric learning. *ICML*, 2007.
[4] R. A. Fisher. The use of multiple measurements in taxonomic problems. *Annual of Eugenics*, 1936.
[5] Y. Freund and R. E. Schapire. A decision-theoretic generalization of on-line learning and an application to boosting. *Journal of Computer and System Sciences*, 55(1):119–139, 1997.
[6] J. Friedma, T. Hastie, and R. Tibshirani. Sparse inverse covariance estimation with the graphical lasso. *Biostatistics*, 1(8):1–10, 2007.
[7] A. Globerson and S. Roweis. Metric learning by collapsing classes. *NIPS*, 2005.
[8] J. Goldberger, S. Roweis, G. Hinton, and R. Salakhutdinov. Neighbourhood components analysis. *NIPS*, 2005.
[9] D. Goldfarb, S. Ma, and K. Schienberg. Fast alternating linearization methods for minimizing the sum of two convex functions. *Technical report, Department of IEOR, Columbia University*, 2009.
[10] R. L. Goldstone and J. Y. Son. Chapter 2: Similarity. *Cambridge Handbook of Thinking and Reasoning*, pages 13–36, 2005.
[11] M. Guillaumin, J. Verbeek, , and C. Schmid. Is that you? metric learning approaches for face identification. *ICCV*, 2009.
[12] S. C. H. Hoi, W. Liu, M. R. Lyu, and W.-Y. Ma. Learning distance metrics with contextual constraints for image retrieval. *CVPR*, 2006.
[13] G. B. Huang, M. Ramesh, T. Berg, and E. Learned-Miller. Labeled faces in the wild: A database for studying face recognition in unconstrained environments. *University of Massachusetts, Amherst, Technical Report 07-49*, 2007.
[14] P. Li, Y. Fu, U. Mohammed, J. Elder, and S. J. Prince. Probabilistic models for inference about identity. *PAMI*, 2011.
[15] W. Liu, S. Ma, D. Tao, J. Liu, and P. Liu. Semi-supervised sparse metric learning using alternating linearization optimization. *KDD*, 2010.
[16] S. G. Mallat and Z. Zhang. Matching pursuits with time-frequency dictionaries. *IEEE Transactions on Signal Processing*, 41(12), 1993.
[17] B. A. Olshausen and D. J. Field. Sparse coding with an overcomplete basis set: A strategy employed by v1. *Vision Research*, 37(23):3311–3325, 1997.
[18] G.-J. Qi, J. Tang, Z.-J. Zha, T.-S. Chua, and H.-J. Zhang. An efficient sparse metric learning in high-dimensional space via $\ell_1$-penalized log-determinant regularization. *ICML*, 2009.
[19] Y. Taigman and L. Wolf. Leveraging billions of faces to overcome performance barriers in unconstrained face recognition. *ArXiv e-prints*, 2011.
[20] Y. Taigman, L. Wolf, and T. Hassner. Multiple one-shots for utilizing class label information. *The British Machine Vision Conference (BMVC)*, 2009.
[21] O. Tuzel, F. Porikli, and P. Meer. Pedestrian detection via classification on riemannian manifolds. *IEEE Transactions on Pattern Analysis and Machine Intelligence*, 30(10), 2008.
[22] K. Q. Weinberger, J. Blitzer, and L. K. Saul. Distance metric learning for large margin nearest neighbor classification. *NIPS*, 2005.
[23] E. P. Xing, A. Y. Ng, M. I. Jordan, and S. Russell. Distance metric learning, with application to clustering with side-information. *NIPS*, 2002.
[24] L. Yang and R. Jin. Distance metric learning: A comprehensive survey. http://www.cse.msu.edu/yangliu1/frame_survey_v2.pdf, 2006.